\ifdefined\ARXIVWRAPPED\else\documentclass[10pt,twocolumn,letterpaper]{article}\fi%
\IfFileExists{arxiv/arxiv-root.tex}{}{}

\usepackage[pagenumbers]{cvpr}
\providecommand{\linenumbers}{}%

\usepackage{caption}

\usepackage{xspace}
\newcommand{\pictura}{Pictura\xspace}
\newcommand{\alberti}{Alberti\xspace}
\newcommand{\argus}{\textsc{Argus}\xspace}
\newcommand{\argusraster}{\textsc{Argus-Raster}\xspace}
\newcommand{\argusplus}{\textsc{Argus+}\xspace}
\newcommand{\simscale}{\scalebox{0.8}{(SimScale)}}

\newcommand{\vectorized}{Vectorized\xspace}
\usepackage[T1,OT1]{fontenc}
\IfFileExists{T1SourceSansThree-TLF.fd}
  {\newcommand{\figfont}{\fontencoding{T1}\fontfamily{SourceSansThree-TLF}\selectfont}}
  {\newcommand{\figfont}{\fontencoding{T1}\fontfamily{SourceSansPro-TLF}\selectfont}}
\usepackage{tikz}
\usetikzlibrary{positioning,arrows.meta,fit,backgrounds,calc,shapes.geometric,patterns,decorations.text}
\definecolor{enccol}{RGB}{35,85,145}
\definecolor{cyclistcol}{RGB}{206,92,0}
\definecolor{trajcol}{RGB}{117,80,123}
\makeatletter
\def\cvprsubsection{\@startsection {subsection}{2}{\z@}
   {-8pt plus -2pt minus -2pt}{5pt} {\fontsize{11}{12}\selectfont\bfseries}}
\def\cvprsubsubsection{\@startsection {subsubsection}{3}{\z@}
   {-6pt plus -2pt minus -2pt}{3pt} {\normalsize\bfseries}}
\makeatother

\usepackage{colortbl}
\usepackage{multirow}
\definecolor{vecrow}{gray}{0.86}
\colorlet{rasterrow}{enccol!14}
\definecolor{keycol}{gray}{0.92}
\definecolor{arguscol}{RGB}{200,95,40}\colorlet{argusrow}{arguscol!18}

\newcommand{\seedfmt}[1]{\pgfmathprintnumber[fixed, precision=1, fixed zerofill, assume math mode=true]{#1}}%
\newcommand{\seedcell}[1]{%
  \gdef\seedn{0}\gdef\seedsum{0}%
  \foreach \v in {#1} {\pgfmathparse{\seedn+1}\xdef\seedn{\pgfmathresult}\pgfmathparse{\seedsum+\v}\xdef\seedsum{\pgfmathresult}}%
  \pgfmathparse{\seedsum/\seedn}\xdef\seedmean{\pgfmathresult}%
  \gdef\seedsq{0}%
  \foreach \v in {#1} {\pgfmathparse{\seedsq+(\v-\seedmean)*(\v-\seedmean)}\xdef\seedsq{\pgfmathresult}}%
  \pgfmathparse{\seedn>1 ? sqrt(\seedsq/max(\seedn-1,1)) : 0}\xdef\seedsd{\pgfmathresult}%
  \pgfmathparse{\seedmean+0.0005}\xdef\seedmean{\pgfmathresult}%
  \pgfmathparse{\seedsd+0.0005}\xdef\seedsd{\pgfmathresult}%
  \seedfmt{\seedmean}\,{\tiny$\pm$\,\seedfmt{\seedsd}}}

\definecolor{linkblue}{rgb}{0.21,0.49,0.74}
\usepackage[breaklinks,colorlinks,allcolors=linkblue]{hyperref}

\usepackage{fontawesome5}
\title{\textsc{Argus}: From Rasterized-Camera Self-Play to Real-Camera Driving%
}

\author{%
Yuan Yin$^{1,*}$ \quad Elias Ramzi$^{1,*}$ \quad Tuan-Hung Vu$^{1}$ \quad Marc Lafon$^{1}$ \quad Yihong Xu$^{1}$ \\
{\'E}loi Zablocki$^{1}$ \quad Alexandre Boulch$^{1}$ \quad Andrei Bursuc$^{1}$ \quad Matthieu Cord$^{1,2}$ \\[6pt]
$^{1}$valeo.ai, Paris, France \quad $^{2}$Sorbonne Universit\'e, CNRS, ISIR, F-75005 Paris, France \\
{\small\texttt{\{firstname.lastname\}@valeo.com}}
}

\newcommand{\projectpage}{\url{https://valeoai.github.io/Argus}}

\begin{document}
\twocolumn[{%
\renewcommand\twocolumn[1][]{#1}%
\maketitle
\begin{center}
\captionsetup{type=figure}
\figfont
\definecolor{arglabel}{RGB}{238,107,38}%
\definecolor{rastlabel}{RGB}{80,150,230}%
\definecolor{rastplan}{RGB}{55,125,200}%
\includegraphics{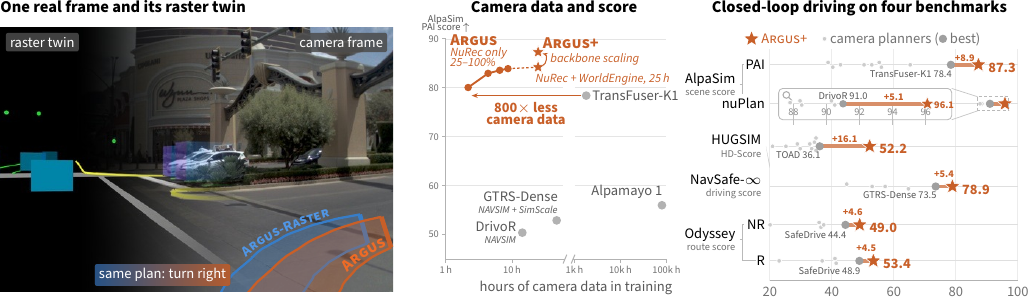}
\DeclareRobustCommand{\argusstar}{\tikz[baseline=-0.6ex]\node[star, star points=5, star point ratio=2.3, fill=arguscol, inner sep=0pt, minimum size=7pt]{};}%
\captionof{figure}{\textbf{Driving learned by reinforcement learning, carried to real cameras.}
  Left: \argusraster drives from the raster twin of a camera frame; aligned to it, \argus reads the camera frame and plans the same right turn.
  Middle: AlpaSim PAI score against the hours of camera data in training; \argus{} (DINOv3 ViT-S/16) aligned in scenes reconstructed from 25 to 100\% of the NuRec logs, and from the NuRec and WorldEngine (navtrain) logs (\protect\argusstar); \argusplus, aligned on the same data from DINOv2 ViT-B/14.
  Right: closed-loop scores of \argusplus and of the imitation-learned camera planners we compare with.
  }
\label{fig:teaser}
\end{center}

}]
{\renewcommand{\thefootnote}{}\footnotetext{$^{*}$Equal contribution.}}%
\addtocontents{toc}{\protect\setcounter{tocdepth}{-1}}
\begin{abstract}
Self-play reinforcement learning (RL) in simulation lets driving scale with compute rather than with logged human data, but learning from scratch takes billions of steps, so observations must be cheap to render.
Rasterized camera views show what driving mostly depends on, the road, lanes, road users, and traffic lights, at a fraction of the cost of realistic images, and through a pretrained image encoder such as DINOv3, a real camera image resembles its raster. 

We present \argus, a camera policy that inherits all of its driving skills from \argusraster, a raster policy trained by self-play, interleaved toward the end of its training with single-agent PPO in logged scenes.
Only its pretrained image encoder is aligned: it learns to match \argusraster's features on a raster twin of the scene, drawn from the perception labels that driving datasets provide.
Its driving comes from 5M hours of RL in simulation, so \argus needs little camera data.
Aligned on only 2.2 hours of reconstructed camera views of driving scenes, it already beats the best camera planner on AlpaSim PAI, a model trained on 1{,}800 hours of driving.
Aligned on 25 hours of such views, \argus outperforms in closed loop on AlpaSim and HUGSIM every camera policy we compare with, learned by imitation or transferred from self-play, by 5.8 points on PAI and 4.9 on nuPlan, and \argusplus, with a larger image backbone, is ahead of every planner it is compared with on AlpaSim, HUGSIM, NavSafe-$\infty$, and Odyssey, with 87.3 on PAI.
Project page: \projectpage.

\end{abstract}

\section{Introduction}
\label{sec:intro}
End-to-end driving policies that run on real cameras mostly learn by imitating logged human driving~\cite{hu2023uniad,chitta2023transfuser,kirby2026drivor}, so they inherit the coverage of the logs, and their errors compound in closed loop~\cite{ross2011dagger}.
Self-play reinforcement learning (RL) offers another route: one policy drives every agent of a simulated scene, learns from rewards alone, and reaches robust driving with no human data~\cite{cusumanotowner2025gigaflow}, so driving improves with compute instead of with logs.
Its simulators, however, are mostly perception-free~\cite{cusumanotowner2025gigaflow,kazemkhani2025gpudrive,cornelisse2025pufferdrive,he2026drivezero}: learning to drive from scratch takes billions of steps, and rendering realistic camera images at each of them is out of reach, so their policies read the simulator state directly, a privileged input that no real sensor provides.
Rasters avoid choosing between costly realistic images and a privileged state: they keep the simulated world simple while showing it through cameras~\cite{yin2026pictura,feng2026rap,rowe2026gigapixel}.
Driving mostly depends on where the road, the lanes, the other road users, and the traffic lights are in each view of the car, much more than on texture, lighting, or weather.
A raster drawn from the cameras of the car shows exactly this with flat colors, and it is cheap enough to render that self-play keeps its scale~\cite{yin2026pictura}.
A raster also shows only what the cameras can see.
A policy that reads the privileged state can rely on agents hidden behind others and on their exact speeds, which no camera policy can recover.
Such gaps between what a teacher and its student observe limit the student~\cite{nguyen2026lead}, whereas a policy that learns from rasters never relies on what the cameras cannot see.

A real car, however, sees camera images, not rasters.
This gap is smaller than it looks, because strong pretrained image encoders already describe an image much as a raster does.
Trained without labels on over a billion images, encoders such as DINOv3~\cite{simeoni2026dinov3} group the pixels of each object, such as a car or a pedestrian, under one consistent semantic description, which changes little with texture, lighting, or weather, the very details that a raster leaves out.
Aligning such an encoder to a policy learned on rasters then only maps one description of the layout onto another, which little data can teach.
The simulator thus does not have to look real: seen through the encoder, a real image already resembles its raster.

\argus builds on this observation in two phases.
First, in simulation, the raster policy \argusraster learns to drive by self-play on rasterized camera views, over 5M hours of driving.
Toward the end of its training, self-play is interleaved with single-agent PPO in logged scenes, which adapts it to the traffic of the logs. A trajectory head, trained last on the frozen policy, then lets it output plans.
Second, an alignment brings it to real cameras.
For each camera view of a real scene, the rasterizer used in self-play renders a raster twin from the same scene state: the HD map, the 3D boxes of the road users, and the traffic-light states (\autoref{fig:teaser}, left).
These are perception labels, which driving datasets already provide, and no trajectory is ever a target.
They can even come from automatic detection and tracking, as in the Waymo Open Motion Dataset~\cite{ettinger2021womd}.
A DINOv3 encoder replaces the raster encoder of \argusraster and is aligned so that the frozen driving head acts on each camera view as it acts on its twin.
The network that learned to drive in simulation is thus the one that drives from real cameras: 25 hours of reconstructed camera views of real scenes teach \argus to see, and all of its driving skills are learned in simulation.

We evaluate \argus in closed loop on four benchmarks: AlpaSim~\cite{nvidia2025alpasim}, on its PAI and nuPlan tracks, HUGSIM~\cite{zhou2026hugsim}, NavSafe-$\infty$~\cite{bao2026navsafe}, which gathers safety-critical scenarios, and Odyssey~\cite{kim2026odyssey}.
\argus needs little camera data: aligned on only 2.2\,h of reconstructed views, it already achieves strong driving performance on the PAI track of AlpaSim (\autoref{fig:teaser}, middle).
Aligned on NuRec and WorldEngine~\cite{li2026worldengine} scenes, it outperforms every imitation-learned camera planner we evaluate, all larger than it, everywhere but on Odyssey in reactive traffic, and \argusplus, aligned from a larger image backbone, outperforms them everywhere (\autoref{fig:teaser}, right).
It also outperforms the prior methods that bring self-play to cameras, whether they distill a privileged teacher into a camera student~\cite{he2026drivezero,dacol2026opted} or align an image encoder to a vectorized privileged policy~\cite{xiong2026terratransfer}.
Finally, analyses and ablations study why it works, from learning on rasters over a vectorized state to the design of its trajectory head and alignment.

Our contributions are:
\begin{itemize}
  \item \argus, a camera driving policy whose driving is learned entirely by RL on rasterized camera views in simulation, in self-play and in logged scenes, and brought to real cameras by aligning a pretrained image encoder to the frozen driving head of \argusraster through raster twins of the same scenes.
\item State-of-the-art closed-loop results among camera policies, ahead of imitation-learned camera planners and of the methods that bring self-play to cameras: on AlpaSim, whose reconstructions provide the alignment views, and on HUGSIM and NavSafe-$\infty$, simulators that the alignment never sees. \argusplus, aligned with the same recipe and data from a larger image backbone, is ahead on all of these and on a third such simulator, Odyssey.
  \item Analyses of why \argus works: a policy learned on rasters transfers to real cameras better than a vectorized one, our recipe gives it a robust trajectory head, and a few hours of reconstructed camera views suffice to align it.
  
\end{itemize}

\section{Related work}
\label{sec:related}

\paragraph{End-to-end driving.}
Most camera driving policies learn by imitating logged human driving.
They range from planners that unify perception and prediction~\cite{hu2023uniad,jiang2023vad} and sensor-fusion transformers~\cite{chitta2023transfuser} to diffusion and scoring-based planners~\cite{liao2025diffusiondrive,li2025gtrs,kirby2026drivor} and vision-language models~\cite{nvidia2025alpamayo}.
Imitation inherits the coverage of the logs, so some planners are refined in closed loop or by search~\cite{li2025ztrs,zhang2026eco}.
Evaluation has moved the same way: open-loop scores on logs reflect closed-loop driving poorly~\cite{dauner2023pdm}, so recent benchmarks drive the policy in pseudo-simulation~\cite{dauner2024navsim} or in real scenes reconstructed by neural rendering, such as HUGSIM~\cite{zhou2026hugsim} and AlpaSim~\cite{nvidia2025alpasim}.
\argus learns to drive from rewards in simulation rather than from human demonstrations, and we evaluate it in closed loop, where its own driving changes what it sees next.

\paragraph{Self-play RL for driving.}
Self-play trains driving policies against copies of themselves in simulation.
Gigaflow~\cite{cusumanotowner2025gigaflow} runs it over 1.6 billion kilometers of simulated driving and reaches state-of-the-art results on three benchmarks with no human data.
GPUDrive~\cite{kazemkhani2025gpudrive} and PufferDrive~\cite{cornelisse2025pufferdrive} make such simulation fast and open, TerraZero~\cite{wu2026terrazero} generates its scenes procedurally, and CaRL~\cite{jaeger2025carl} scales RL planners with a simple route-completion reward.
Human-regularized self-play adds a small amount of human driving data, so that self-play policies become compatible with human traffic~\cite{cornelisse2024humancompatible,cornelisse2026humanlike,chang2025spacer}.
All these policies observe a privileged state.
\pictura~\cite{yin2026pictura} adds a rasterizer to PufferDrive and trains the first large-scale self-play policy directly from rendered camera views, which it evaluates in simulation.
\argusraster follows this idea, with rasters adapted to ease the alignment to real images.

\begin{table}[t]
  \centering
  \caption{\textbf{Design choices of methods that transfer the driving behavior of a self-play policy to a camera policy.}
    Vector: privileged vectorized state; Raster: rasterized camera views.}
  \label{tab:positioning}
  \footnotesize
  \setlength{\tabcolsep}{1.6pt}
  \resizebox{\columnwidth}{!}{%
  \begin{tabular}{@{}lcccc c@{}}
    \toprule
    \multirow{2}{*}[-0.5ex]{Method} & \multicolumn{2}{c}{RL policy} & \multirow{2}{*}[-0.5ex]{Transfer} & \multicolumn{2}{c}{Deployed} \\
    \cmidrule(lr){2-3} \cmidrule(lr){5-6}
    & Input & Output & & Network & Output \\
    \midrule
    Gigapixel~\cite{rowe2026gigapixel} & Vector & Action & DAgger+align. & Student & Trajectory \\
    DriveZero~\cite{he2026drivezero} & Vector & Action & Distillation & Student & Trajectory \\
    OPTED~\cite{dacol2026opted} & Vector & Action & DAgger & Student & Trajectory \\
    TerraTransfer~\cite{xiong2026terratransfer} & Vector & Action & Alignment & Aligned policy & Action \\
    \midrule
    \argus & Raster & Act.\,\&\,Traj. & Alignment & Aligned policy & Act.\,\&\,Traj. \\
    \bottomrule
  \end{tabular}}
\end{table}
\paragraph{From privileged teachers to cameras.}
Learning by Cheating~\cite{chen2019lbc} set the template: a teacher with privileged inputs supervises a sensorimotor student.
Recent methods take an RL teacher on a vectorized state and distill it into a separate camera planner (\autoref{tab:positioning}), offline~\cite{he2026drivezero} or with DAgger~\cite{ross2011dagger}, in a box-world renderer~\cite{rowe2026gigapixel} or a neural simulator~\cite{dacol2026opted}.
The camera planner thus learns to drive a second time, by imitation, and LEAD~\cite{nguyen2026lead} shows that the asymmetries between learner and expert limit such students.
Gigapixel~\cite{rowe2026gigapixel} obtains each DAgger label by forking the simulator at the state that its student reached and rolling the teacher out from there, whereas our trajectory head collects its labels online during self-play, from the rollout itself.
TerraTransfer~\cite{xiong2026terratransfer} instead keeps the frozen head of its vectorized teacher and trains image encoders on (image, state) pairs from driving logs.
These encoders are asked to reproduce a latent of the privileged state, which is unlike an image and can hold what a camera cannot see, such as occluded agents and exact speeds~\cite{nguyen2026lead}.
\argus also deploys the network learned in simulation, but \argusraster already sees camera views rendered in simulation, so the transfer to real cameras stays between two camera views of the same scene.
\autoref{sec:exp:vectorized} tests this difference directly, with a vectorized policy aligned by the same recipe.

\paragraph{Bridging simulated and real images.}
A driving policy trained in simulation on semantic segmentation transfers to a real vehicle once a perception module predicts it from real images~\cite{muller2018driving}.
Self-supervised encoders such as DINOv3~\cite{simeoni2026dinov3} learn such semantics without labels, and \argus uses one as this bridge.
Unpaired domain adaptation aligns feature distributions~\cite{ganin2015dann}, and relational distillation matches relations between samples~\cite{park2019rkd}, as the structural loss of TerraTransfer~\cite{xiong2026terratransfer} does; with paired data, RAP~\cite{feng2026rap} and Gigapixel~\cite{rowe2026gigapixel} align the scene tokens of one backbone across real and rendered frames, both to train a camera planner by imitation.
In \argus, only \argusraster reads the raster twin, during alignment, and the deployed policy drives from camera images.
\argus matches the latents of a large image backbone to those of the raster encoder of \argusraster, and distills its policy~\cite{rusu2016policy}.

\section{\argus: from simulation to real cameras}
\label{sec:method}

\begin{figure*}[t]
  \centering
    \DeclareRobustCommand{\stepno}[3]{\tikz[baseline=(n.base)]\node[circle, fill=#2, text=#3, inner sep=0pt, minimum size=1.25em, font=\footnotesize\bfseries\upshape, line width=0pt] (n) {#1};}%
  \resizebox{\linewidth}{!}{%
\figfont
\includegraphics{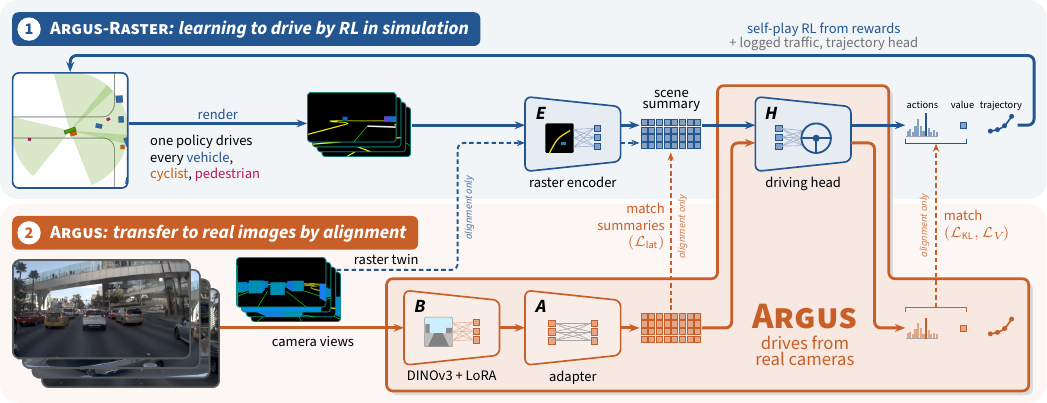}}
  \caption{\textbf{Self-play teaches driving; alignment teaches seeing.}
    \stepno{1}{enccol}{white}~\argusraster learns to drive by self-play RL on rasterized camera views, one policy driving every agent of the scene; an extension of self-play to logged traffic and a trajectory head then complete it.
    \stepno{2}{arguscol}{white}~To drive from real cameras, only its encoder is replaced.
    For each camera view, the frozen \argusraster reads a raster twin of the same scene.
    A new encoder $E'$ of an image backbone $B$ and an adapter $A$ learns from the camera view to match the scene latents of \argusraster ($\mathcal{L}_\text{lat}$), and the distribution over actions ($\mathcal{L}_\text{KL}$) and value ($\mathcal{L}_V$) that the frozen head $H$ gives from it.
    Dashed lines are used only during alignment; at deployment, \argus reads only real camera frames.
    The trajectory head is trained only in simulation, used zero-shot.}
  \label{fig:overview}
\end{figure*}

\argus is built in two phases (\autoref{fig:overview}).
It first learns to drive in a raster world, where copies of \argusraster drive every agent and learn from one another's mistakes; \autoref{sec:method:selfplay} describes this phase.
Only then does it learn to see: \autoref{sec:method:alignment} gives it an encoder of real images and keeps its driving head frozen, so everything that RL taught it in closed loop carries over to real cameras in the camera policy \argus.
By default, both drive with their trajectory head.

\subsection{\argusraster: learning in a raster world}
\label{sec:method:selfplay}

\paragraph{A simulator that renders cameras.}
Driving learned in simulation is more likely to carry over to real cameras if the policy already sees the world as a camera does.
\argusraster therefore learns in a fast driving simulator built on PufferDrive~\cite{cornelisse2025pufferdrive}.
\emph{Cameras.} As in \pictura~\cite{yin2026pictura}, the simulator renders the camera views of every controlled agent at every step.
Each vehicle draws its camera rig from a pool at the start of an episode, and the renderer supports both pinhole and lens cameras.
\emph{Views.} The views show flat-shaded road users, lanes, road markings, and traffic lights, with the perspective and the occlusions of a camera.
Elements with no physical counterpart in an image are dropped, and buildings are painted black so that they only occlude.
\emph{Scenes.} The maps are road networks converted to the simulator.
Their scenes range from empty roads to locally dense traffic of up to 300 road users.
Pedestrians cross the road, and parked cars line the curbs.
Agents move as kinematic bicycles driven through jerk, as in Gigaflow~\cite{cusumanotowner2025gigaflow} (\autoref{sec:impl:sim}).

\paragraph{Self-play at scale.}
\label{sec:method:training}
In this simulator, copies of \argusraster drive every agent of a scene: vehicles, cyclists, and pedestrians.
\argusraster learns by PPO~\cite{schulman2017ppo} from the rewards that all of them collect.
Since every agent is a learner, one run gathers billions of agent steps of closed-loop interaction without a single human label.
The rewards follow Gigaflow~\cite{cusumanotowner2025gigaflow}.
Reaching the waypoints of the route is rewarded.
Collisions, leaving the road, and running red lights and stop signs are penalized.
Shaping terms reward progress and lane keeping and penalize discomfort and overspeed (see \autoref{sec:impl:selfplay} for the rewards).
At every spawn, each agent also draws its own reward coefficients, dynamics multipliers, and speed cap.
A single policy thus learns a whole population of driving styles and vehicles.

\paragraph{One driving head shared across camera rigs.}
Each agent observes a single frame of the rasterized views $\mathbf{x}^\text{ras}$ of its cameras.
It also observes its ego state and a conditioning vector of its goals and of what it drew at spawn; we write $\mathbf{s}$ for the two together.
As in Gigaflow~\cite{cusumanotowner2025gigaflow}, it chooses an action $a$ from the distribution $\pi(a \mid \mathbf{x}^\text{ras}, \mathbf{s})$ over a set $\mathcal{A}$ of 12 jerks.
Each action pairs a longitudinal jerk in $\{-15,-4,0,4\}$\,m/s$^3$ with a lateral jerk in $\{-4,0,4\}$\,m/s$^3$.
PPO also learns a critic that estimates the value $V^\pi(\mathbf{x}^\text{ras}, \mathbf{s})$, the return expected from this state under $\pi$.
To make later deployment flexible, the network is designed so that its driving part is shared across camera rigs.
The network therefore splits into an encoder $E$ of the raster views and a head $H$.
The head maps the output of $E$ and $\mathbf{s}$ to both the distribution over $\mathcal{A}$ and the value:
\begin{equation}
  \big(\pi(\cdot \mid \mathbf{x}^\text{ras}, \mathbf{s}),\, V^\pi(\mathbf{x}^\text{ras}, \mathbf{s})\big) = H\big(E(\mathbf{x}^\text{ras}), \mathbf{s}\big).
  \label{eq:policy}
\end{equation}
The encoder $E$ works in two transformer stages.
A raster backbone $C$ first summarizes each of the $K$ cameras into a few tokens.
It also receives the viewing direction of every image patch, so the same weights serve different cameras.
A scene pooler $P$ then gathers the tokens of all cameras into a fixed number of scene latents: $E(\mathbf{x}^\text{ras}) = P\big(\{C(\mathbf{x}^\text{ras}_k)\}_{k=1}^{K}\big)$.
This pooler is a set encoder over the cameras.
Whatever the rig, the head thus receives the same latents.
The rest of \argus builds on this property.
One head drives with every rig of the self-play pool, and the alignment only has to reproduce these latents, whatever the rig of each log.
In $H$, a transformer trunk combines the scene latents with $\mathbf{s}$.
Two linear heads read its learned output tokens: one gives the distribution over the 12 actions, the other $V^\pi$.
Of these two heads, the action head drives; the trajectory head added in \autoref{sec:method:traj} drives by default.
\autoref{fig:architecture_detail} in \autoref{sec:impl:arch} shows the architecture.

\paragraph{Extending self-play to logged traffic.}
\label{sec:method:logged}
Self-play traffic is not human traffic.
Moreover, log-replay driving tasks replay traffic that does not react to the ego~\cite{cornelisse2025reliable,distelzweig2026beyond}.
Self-play is therefore extended with a second world, in the same PPO training: in every batch, its own self-play world runs side by side with logged scenes that are converted to the simulator and rasterized.
In the logged scenes, it drives the ego while every other agent replays its log, as in the non-reactive traffic in which CaRL trains its PPO planner~\cite{jaeger2025carl}.
Its goals lie on the logged route when the log provides one, and its reward can be adapted to the task's driving rules.
The log sets the scene and the traffic, but the logged trajectory of the ego is never a training target.
\argusraster thus learns to drive among humans without imitating one.
In its own world, it keeps driving every agent, so it retains its skills.

\paragraph{Trajectory head by online self-distillation.}
\label{sec:method:traj}
Many driving stacks expect a planned trajectory, whereas self-play systems output one action per step~\cite{cusumanotowner2025gigaflow,cornelisse2025pufferdrive,he2026drivezero,wu2026terrazero}.
Rather than learning to plan by RL from scratch, \argusraster teaches itself.
A trajectory head is added to the frozen network and learns online, in the rollouts of the policy, to predict the paths that the action head drives.
This head generates a plan by flow matching~\cite{liu2023rectifiedflow} on top of the frozen trunk.
It reads the scene latents, the action read-out of the trunk, and the action that the action head chooses.
From them, it predicts the short plan that the action head then drives.
When the trajectory head drives, a controller tracks this plan.
All other weights are shared with the action head, which stays unchanged.
To train it, the two heads take turns driving: the action head drives a segment, then hands over to the trajectory head for the next one.
The trajectory head learns from the states where the action head drives.
Its target is the path that the action head then drives until the next handover.
As in DAgger~\cite{ross2011dagger}, the trajectory head thus drives during its own training.
Each segment of the action head starts from a state that the trajectory head reached, so its targets teach the trajectory head to recover from its own errors.
With labels from the rollout itself, training needs no separate expert rollout, no copy of the simulator state, and no replay buffer.

\subsection{\argus: alignment to camera images}
\label{sec:method:alignment}
\begin{table*}[t]
  \centering
  \renewcommand{\simscale}{\unskip\textsuperscript{$\diamond$}}%
  \caption{\textbf{Closed-loop results on AlpaSim, NavSafe-$\infty$ and Odyssey.}
    (a)~AlpaSim, PAI and nuPlan tracks. Score: official AlpaSim metric, route progress capped at 80\%, zero for a failed scene; Prog.: progress along the logged route, clamped to [0, 1] per scene (\%); km/inf.: km per at-fault collision or off-road event; full metrics in \autoref{tab:alpasim_all}.
    Each baseline with the better of AlpaSim's two MPCs; \textbf{bold}, \underline{underlined}: best, second best.
    (b)~NavSafe-$\infty$: driving score and success rate. (c)~Odyssey: route driving score in non-reactive and reactive traffic. Numbers of (b, c) in \autoref{tab:navsafe_odyssey}.
}
  \label{tab:alpasim}
  \vspace{2pt}%
  \begin{minipage}[t]{0.4245\textwidth}\vspace{0pt}%
    {\centering\scriptsize\bfseries (a) AlpaSim, PAI and nuPlan tracks\par}%
    \vspace{2pt}%
      \centering
 
  \scriptsize
  \setlength{\tabcolsep}{2.5pt}
  \renewcommand{\arraystretch}{1.07}%
  \resizebox{\linewidth}{!}{%
  \begin{tabular}{lr>{\columncolor{keycol}}ccc>{\columncolor{keycol}}ccc}
    \toprule
    & & \multicolumn{3}{c}{PAI track} & \multicolumn{3}{c}{nuPlan track} \\
    \cmidrule(lr){3-5} \cmidrule(lr){6-8}
    Method & Params. & Score\,$\uparrow$ & Prog.\,$\uparrow$ & km/inf.\,$\uparrow$ & Score\,$\uparrow$ & Prog.\,$\uparrow$ & km/inf.\,$\uparrow$ \\
    \midrule
    Qwen-Drive 1.0~\cite{zhou2026qwendrive} & 5.1B & 25.4$^{\ddagger}$ & 39.1 & 0.21 & -- & -- & -- \\
    LTFv6~\cite{nguyen2026lead} & 62.1M & 26.5$^{\ddagger}$ & 43.7 & 0.23 & -- & -- & -- \\
    LTFv6 + OPTED~\cite{dacol2026opted} & 62.1M & 41.8$^{\ddagger}$ & 73.4 & 0.43 & -- & -- & -- \\
    Alpamayo 1~\cite{nvidia2025alpamayo} & 10.5B & 56.0$^{\dagger}$ & 75.3 & 0.96 & -- & -- & -- \\
    Alpamayo 1.5~\cite{nvidia2025alpamayo} & 10.5B & 55.3$^{\dagger}$ & 76.3 & 0.94 & -- & -- & -- \\
    LTF \simscale~\cite{chitta2023transfuser,tian2026simscale} & 56.0M & 38.8 & 50.4 & 0.41 & 88.4 & 89.3 & 0.43 \\
    DiffusionDrive \simscale~\cite{liao2025diffusiondrive,tian2026simscale} & 60.8M & 40.3 & 44.6 & 0.49 & 87.8 & 88.8 & 0.43 \\
    GTRS-Dense \simscale~\cite{li2025gtrs,tian2026simscale} & 81.0M & 52.9 & 66.5 & 0.66 & 88.5$^{\dagger}$ & 82.9 & 0.35 \\
    DrivoR~\cite{kirby2026drivor} & 40.8M & 50.4 & 53.2 & 0.68 & 91.0 & 90.9 & 0.63 \\
    DrivoR \simscale~\cite{kirby2026drivor,tian2026simscale} & 40.8M & 42.8 & 41.8 & 0.91 & 90.5 & 87.9 & 1.06 \\
    TransFuser-K1~\cite{kesai2026py123dgarage} & 67.7M & 78.4$^{\dagger}$ & 84.6 & 1.44 & 90.3$^{\dagger}$ & 79.6 & 0.72 \\
    \rowcolor{argusrow} \argus & 31.7M & \underline{84.2} & \underline{90.6} & \underline{1.90} & \underline{95.9} & \underline{94.9} & \underline{1.72} \\
    \rowcolor{argusrow} \argusplus & 98.5M & \textbf{87.3} & \textbf{91.9} & \textbf{2.84} & \textbf{96.1} & \textbf{95.0} & \textbf{2.00} \\
    \bottomrule
  \end{tabular}}

    {\scriptsize\raggedright\noindent$^{\dagger}$\,nonlinear MPC, \texttt{py123d\_garage}~\cite{kesai2026py123dgarage};\enspace$^{\ddagger}$\,as reported by OPTED~\cite{dacol2026opted}, scored before AlpaSim's corridor fix (see \autoref{sec:protocol:compare});\enspace$^{\diamond}$\,SimScale checkpoint~\cite{tian2026simscale}\par}%
  \end{minipage}\hfill
  \begin{minipage}[t]{0.5605\textwidth}\vspace{0pt}
      \centering
  \resizebox{\linewidth}{!}{%
\figfont
\includegraphics{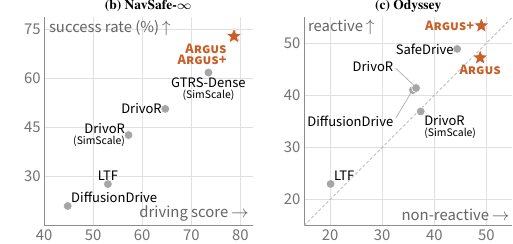}}

    {\centering\parbox{0.92\linewidth}{\scriptsize\raggedright Baselines as published, collected by \texttt{py123d\_garage}~\cite{kesai2026py123dgarage}. Our runs: \mbox{DrivoR \scalebox{0.8}{(SimScale)}}, \argus{} and \argusplus{} (mean of 3 seeds on NavSafe-$\infty$).}\par}%
  \end{minipage}
\end{table*}

\argusraster learns to drive from raster views, while a real car records camera images.
The alignment brings it to these images without changing how it drives: it trains only a new encoder, and the driving head that RL optimized in closed loop stays frozen.
The encoder learns from pairs of a camera frame and a raster view of the same scene.

\paragraph{Raster twins.}
Driving logs already contain what the rasterizer of the simulator needs to create the raster views: 3D boxes, traffic-light states, and an HD map, next to the camera images.
For each frame, this rasterizer draws the map, the road users, and the traffic lights through the cameras of that frame.
The result is the raster twin $\mathbf{x}^\text{ras}$ of the real frame, with $\mathbf{s}$ filled from the log.
Its encoder $E$ gives the target latents $E(\mathbf{x}^\text{ras})$.
The twin inherits the limits of the logs and the map, such as errors in the logged boxes and light states. In our experiments, the camera views are reconstructed by AlpaSim from logged scenes, along drives of \argusraster; these scenes give no light states, so the lights are drawn unlit, and the twin is seen through the closest PAI rig or the nuPlan rig (\autoref{sec:impl:align}).

\paragraph{An image encoder that mirrors the raster one.}
The camera policy replaces the raster encoder $E$ with an image encoder $E'$ of camera images $\mathbf{x}^\text{cam}$ and keeps the driving head $H$ of \autoref{eq:policy} frozen.
$E'$ mirrors the two stages of $E$, so that each stage has a raster counterpart.
An image backbone $B$ takes the place of the raster backbone $C$.
An adapter $A$, also a set encoder over the cameras, takes the place of the scene pooler $P$.
Thus $E'(\mathbf{x}^\text{cam}) = A\big(\{B(\mathbf{x}^\text{cam}_k)\}_{k=1}^{K}\big)$, and the frozen head gives the camera policy and its value:
\begin{equation}
  \big(\pi_\text{cam}(\cdot \mid \mathbf{x}^\text{cam}, \mathbf{s}),\, V_\text{cam}(\mathbf{x}^\text{cam}, \mathbf{s})\big) = H\big(E'(\mathbf{x}^\text{cam}), \mathbf{s}\big).
  \label{eq:camera_policy}
\end{equation}
The backbone $B$ is a pretrained vision transformer adapted with LoRA~\cite{hu2022lora}, so the camera policy starts from the visual knowledge of large-scale pretraining.
Learned scene tokens are added to its input, and their outputs are the scene tokens of the camera.
The adapter $A$ maps them to the summary tokens that $C$ gives for that camera.
Its last part is a trainable copy of the scene pooler $P$, which gathers the tokens of all cameras into the latents that $H$ reads.
\autoref{sec:impl:align} details the LoRA and the adapter.

\paragraph{Losses.}
The image encoder $E'$ is trained with three standard losses for aligning a student to a frozen policy.
The latent loss $\mathcal{L}_\text{lat}$ regresses the camera latents $E'(\mathbf{x}^\text{cam})$ onto the raster latents $E(\mathbf{x}^\text{ras})$, as RAP and Gigapixel align the scene tokens of real and rendered frames~\cite{feng2026rap,rowe2026gigapixel}.
The policy loss $\mathcal{L}_\text{KL}$ is the KL divergence between the action distributions $\pi(\cdot \mid \mathbf{x}^\text{ras}, \mathbf{s})$ and $\pi_\text{cam}(\cdot \mid \mathbf{x}^\text{cam}, \mathbf{s})$, as in policy distillation~\cite{rusu2016policy}.
The value loss $\mathcal{L}_V$ regresses $V_\text{cam}$ onto $V^\pi$, as in ITER~\cite{igl2021transient}.
Their sum is the loss of \argus, $\mathcal{L}_\text{\argus} = \mathcal{L}_\text{lat} + \mathcal{L}_\text{KL} + \mathcal{L}_V$.
\autoref{sec:impl:align} details them.

\section{Experimental results}
\label{sec:exp}

The experiments answer three questions.
Does a policy that learned to drive by RL on raster views~\cite{yin2026pictura} drive better from real cameras than the camera planners trained on logged driving (\autoref{sec:exp:main})?
What do self-play, its extension to logged traffic, and the trajectory head bring to \argusraster (\autoref{sec:exp:pipeline})?
How much of its driving does the alignment keep, and does a raster policy keep more than a vectorized one (\autoref{sec:exp:vectorized})?
\autoref{sec:exp:setting} gives the setup, and \autoref{sec:exp:ablation} the ablations.

\subsection{Setup}
\label{sec:exp:setting}

\noindent\textbf{Learning in simulation.}
\argusraster has 4.9M parameters, 2.9M without its trajectory head; its raster encoder turns the $96{\times}54$ camera rasters into 32 latents, read by its driving head.
It trains by self-play on 8 CARLA towns~\cite{dosovitskiy2017carla} and 4 nuPlan cities~\cite{caesar2021nuplan}, with camera rigs drawn from a pool of ten.
Self-play runs PPO for about 60B agent steps of 0.3\,s on 260 A100 GPUs, which simulate about 1.33M agents in parallel, for 2.1k A100 GPU-hours.
The self-play of \argusraster is then extended to logged traffic: a quarter of its environments keep running its self-play world, and the others replay 1{,}458 of the 1{,}761 training scenes of the PAI track and the 12{,}861 scenes of NAVSIM navtrain, the training split of the nuPlan track, as AlpaSim runs them.
This one \argusraster serves every benchmark.
The run lasts 342M steps, a quarter of them in the self-play world, for 32 H100 GPU-hours.
Its trajectory head, which plans 3\,s ahead as 10 poses 0.3\,s apart, is trained last, on the frozen network in these logged scenes, and its probability of driving after a handover (\autoref{sec:method:traj}) is linearly warmed up to 1, for 32 H100 GPU-hours as well.

\vspace{0.4em}%
\noindent\textbf{Alignment.}
The backbone of \argus is a DINOv3 ViT-S/16~\cite{simeoni2026dinov3} with LoRA of rank 32, and its adapter maps the backbone tokens to the 32 latents of the raster encoder (\autoref{sec:method:alignment}).
A single \argus is aligned from this \argusraster for every benchmark, on the reconstructed camera frames that AlpaSim produces along its drives in the training scenes of its two tracks, NuRec reconstructions of PAI and WorldEngine~\cite{li2026worldengine} reconstructions of NAVSIM navtrain~\cite{dauner2024navsim}.
Keeping only the drives with an AlpaSim score of 1 (\autoref{sec:protocol:benchmarks}; studied in \autoref{sec:ablation:data}) leaves 25\,h of driving, 8.7\,h on NuRec and 16.1\,h on WorldEngine (\autoref{sec:impl:align_data}). The alignment takes 48 H100 GPU-hours, about 86 A100 GPU-hours (\autoref{tab:backbone_gpuh}).
We also align \argusplus with the same recipe and data from a larger backbone, DINOv2 ViT-B/14~\cite{oquab2024dinov2}, for 328 A100 GPU-hours (\autoref{tab:backbone_gpuh}).

\vspace{0.4em}%
\noindent\textbf{Benchmarks.}
\phantomsection\label{sec:exp:bench}
We evaluate in closed loop on the PAI and nuPlan tracks of AlpaSim~\cite{nvidia2025alpasim}, on NavSafe-$\infty$~\cite{bao2026navsafe} and Odyssey~\cite{kim2026odyssey} (panels (a), (b), and (c) of \autoref{tab:alpasim}), and on HUGSIM~\cite{zhou2026hugsim} (\autoref{tab:hugsim}), all in their default settings and with their official scores; HUGSIM's official HD-Score scores the trajectory that a policy plans.
On HUGSIM, we also report the realized HD-Score of TerraTransfer~\cite{xiong2026terratransfer}, which scores the path that the ego drove instead, and its non-default 10\,Hz setting, which matches the 0.1\,s step of our policies.
On AlpaSim and HUGSIM, our policies drive with their trajectory head unless marked w/~action, and with one fixed setting of their conditioning (\autoref{sec:impl:selfplay}).

\vspace{0.4em}%
\noindent\textbf{Baselines.}
\phantomsection\label{sec:exp:baselines}
The baselines fall into two groups.
\emph{Camera planners} (\cref{tab:alpasim,tab:hugsim}).
Imitation-learned planners~\cite{hu2023uniad,jiang2023vad,chitta2023transfuser,nguyen2026lead,kesai2026py123dgarage,liao2025diffusiondrive,li2025gtrs,kirby2026drivor,xu2026toad,bartoccioni2025vavam,zhou2026qwendrive,dacol2026opted}, vision-language models~\cite{nvidia2025alpamayo}, and planners refined in closed loop or by search~\cite{li2025ztrs,zhang2026eco}.
Several planners also have a checkpoint trained on the synthetic data of SimScale~\cite{tian2026simscale}, marked (SimScale).
Unless marked as reported by another work (\autoref{sec:protocol:compare}), each row is our run of the released checkpoint, and one closed-loop refinement~\cite{zhang2026eco} is our reimplementation.
\emph{RL teachers distilled to cameras} (\autoref{tab:hugsim_bc}).
Gigapixel~\cite{rowe2026gigapixel} and DriveZero~\cite{he2026drivezero} release no camera policy, and TerraTransfer~\cite{xiong2026terratransfer} evaluates under its own protocol; we compare with the results these papers publish, for themselves and for the planners they evaluate.

See \autoref{sec:impl} for the implementation of each stage and \autoref{sec:protocol} for the protocol of each benchmark and baseline.

\subsection{Comparison with camera planners}
\label{sec:exp:main}

\begin{table*}[t]
  \centering
  \caption{\textbf{Closed-loop results on HUGSIM.}
    (a)~Scores split by the dataset each scenario comes from: nuScenes~\cite{caesar2020nuscenes}, PandaSet~\cite{xiao2021pandaset}, KITTI-360~\cite{liao2023kitti360}, and Waymo~\cite{sun2020waymo}; \textbf{bold}, \underline{underlined}: best, second best.
    (b)~The protocol of Gigapixel and DriveZero. (c)~The protocol of TerraTransfer.
    See \autoref{sec:more_results:compare} for the detailed scores.}
  \label{tab:hugsim}\label{tab:hugsim_bc}
  \vspace{2pt}%
  \begin{minipage}[t]{0.62\textwidth}\vspace{0pt}%
    {\centering\scriptsize\bfseries (a) Default setting, 436 official scenarios, by source dataset\par}%
    \vspace{2pt}%
    
  \centering
  \footnotesize
  \setlength{\tabcolsep}{2pt}
  \resizebox{\linewidth}{!}{%
  \begin{tabular}{lr>{\columncolor{keycol}}ccccc>{\columncolor{keycol}}ccccc}
    \toprule
    & & \multicolumn{5}{c}{HD-Score $\uparrow$} & \multicolumn{5}{c}{Realized HD-Score $\uparrow$} \\
    \cmidrule(lr){3-7} \cmidrule(l){8-12}
    Method & Params. & All & nuSc. & Panda. & KITTI & Waymo & All& nuSc. & Panda. & KITTI & Waymo \\
    \midrule
     UniAD~\cite{hu2023uniad} & 131.8M & 29.8 & 34.3 & 38.4 & 6.3 & 40.4 & 39.4& 45.3 & 45.5 & 17.1 & 50.9 \\
     VAD~\cite{jiang2023vad} & 58.4M & 12.4 & 17.2 & 15.8 & 3.9 & 13.3 & 23.3& 30.1 & 29.0 & 9.1 & 25.8 \\
     LTF~\cite{chitta2023transfuser} & 56.0M & 20.9 & 30.5 & 27.0 & 7.5 & 20.1 & 30.6& 41.0 & 39.6 & 13.0 & 29.9 \\
     LTFv6~\cite{nguyen2026lead} & 62.1M & 34.9 & 42.3 & 38.8 & 23.2 & 36.5 & 41.3& 49.5 & 45.5 & 27.4 & 44.4 \\
     GTRS-Dense~\cite{li2025gtrs} & 81.0M & 34.4 & 42.9 & 33.9 & 25.4 & 37.4 & 41.1& 50.1 & 43.7 & 28.1 & 44.1 \\
     GTRS-Aug~\cite{li2025gtrs} & 86.4M & 28.9 & 37.0 & 32.4 & 20.2 & 27.3 & 36.1& 44.2 & 42.7 & 23.8 & 34.7 \\
     DrivoR~\cite{kirby2026drivor} & 40.8M & 33.5 & 37.9 & 35.8 & 22.0 & 39.0 & 40.1& 43.4 & 45.8 & 25.3 & 46.0 \\
     DrivoR + ECO$^{*}$~\cite{zhang2026eco} & 40.8M & 32.9 & 37.5 & 34.7 & 19.5 & 41.0 & 39.2& 43.0 & 44.6 & 22.4 & 47.4 \\
     TOAD~\cite{xu2026toad} & 40.8M & 36.1 & 44.3 & 36.4 & 23.0 & 42.6 & 41.8& 49.4 & 46.2 & 25.4 & 47.4 \\
     ZTRS~\cite{li2025ztrs} & 83.4M & 34.9 & 42.2 & 34.9 & 29.1 & 35.0 & 42.0& 48.6 & 45.8 & 32.0 & 42.5 \\
     VaVAM-B~\cite{bartoccioni2025vavam} & 357.3M & 13.4 & 20.3 & 16.9 & 8.0 & 9.2 & 29.3& 40.7 & 36.6 & 19.8 & 21.5 \\
     VaVAM-B + ECO$^{*}$~\cite{zhang2026eco} & 357.3M & 25.2 & 35.3 & 33.5 & 16.4 & 16.3 & 32.2& 47.0 & 39.6 & 22.1 & 22.1 \\
    \rowcolor{argusrow} \argus & 31.7M & \textbf{53.6} & \underline{52.9} & \textbf{57.0} & \textbf{33.4} & \textbf{71.2} & \textbf{57.4} & \underline{60.9} & \textbf{60.3} & \textbf{36.6} & \textbf{73.0} \\
    \rowcolor{argusrow} \argusplus & 98.5M & \underline{52.2} & \textbf{57.0} & \underline{55.3} & \underline{31.7} & \underline{66.2} & \underline{55.6} & \textbf{63.8} & \underline{58.3} & \underline{34.7} & \underline{67.7} \\
    \bottomrule
  \end{tabular}}

    {\scriptsize\raggedright\noindent$^{*}$\,our reimplementation\par}\vspace{-8pt}%
  \end{minipage}\hfill
  \begin{minipage}[t]{0.365\textwidth}\vspace{0pt}
      \centering
  \resizebox{\linewidth}{!}{%
\figfont
\begin{tikzpicture}[font=\scriptsize, x=0.094cm, y=0.2716cm]%
\tikzset{
  cam/.style={fill=black!20, draw=none},
  arg/.style={fill=arguscol, draw=arguscol},
  argplus/.style={arg},
}
\newcommand{\pbar}[7]{%
  \path[#5] (0,{-#1-0.4}) -- (0,{-#1+0.4}) [rounded corners=1.5pt] -- (#4,{-#1+0.4}) -- (#4,{-#1-0.4}) [sharp corners] -- cycle;%
  \node[anchor=west, inner sep=1.2pt, font=\scriptsize, text=#7] at (0,-#1) {#2\,{\tiny\color{black!60}#3}};%
  \node[anchor=west, inner sep=1.2pt, font=\scriptsize] at (#4,-#1) {#6};}
\newcommand{\zeroaxis}[2]{\draw[black!45, line width=0.4pt] (0,{-#1+0.6}) -- (0,{-#2-0.55});}
\newcommand{\panel}[3]{%
  \foreach \g in {0,20,40,60} {\draw[black!12, line width=0.3pt] (\g,{-#1+0.6}) -- (\g,{-#2-0.55});
    \node[font=\scriptsize, text=black!55, anchor=north, inner sep=1pt] at (\g,{-#2-0.55}) {\g};}
  \node[anchor=south, inner sep=0pt, font=\rmfamily\scriptsize\bfseries] at (30,{-#1+0.75}) {#3};}%
\panel{1}{9}{(b) Pre-challenge set, 345 scenarios, official HD-Score}
\pbar{1}{\textbf{\argus}}{}{54.1}{arg}{\textbf{54.1}}{white}%
\pbar{2}{\textbf{\argusplus}}{}{54.3}{argplus}{\textbf{54.3}}{white}
\pbar{3}{DriveZero-Scale}{\cite{he2026drivezero}}{46.6}{cam}{46.6}{black}
\pbar{4}{DriveZero}{\cite{he2026drivezero}}{39.4}{cam}{39.4}{black}
\pbar{5}{Gigapixel}{\cite{rowe2026gigapixel}}{38.5}{cam}{38.5}{black}
\pbar{6}{DrivoR}{\cite{kirby2026drivor}}{35.7}{cam}{35.7}{black}
\pbar{7}{UniAD}{\cite{hu2023uniad}}{32.7}{cam}{32.7}{black}
\pbar{8}{LTF}{\cite{chitta2023transfuser}}{23.7}{cam}{23.7}{black}
\pbar{9}{VAD}{\cite{jiang2023vad}}{13.4}{cam}{13.4}{black}
\zeroaxis{1}{9}
\begin{scope}[yshift=-3.3636cm]%
\panel{1}{7}{(c) nuScenes-10\,Hz, 88 scenarios, realized HD-Score}
\pbar{1}{\textbf{\argus}}{}{62.1}{arg}{\textbf{62.1}}{white}%
\pbar{2}{\textbf{\argusplus}}{}{60.7}{argplus}{\textbf{60.7}}{white}
\pbar{3}{TerraTransfer}{\cite{xiong2026terratransfer}}{49.0}{cam}{49.0}{black}
\pbar{4}{ECO}{\cite{zhang2026eco}}{45.2}{cam}{45.2}{black}
\pbar{5}{LTF}{\cite{chitta2023transfuser}}{36.0}{cam}{36.0}{black}
\pbar{6}{VAD}{\cite{jiang2023vad}}{23.9}{cam}{23.9}{black}
\pbar{7}{UniAD}{\cite{hu2023uniad}}{22.4}{cam}{22.4}{black}
\zeroaxis{1}{7}
\end{scope}
\end{tikzpicture}}
  \end{minipage}
\end{table*}
\noindent\textbf{\argus outperforms imitation-learned camera planners.}
\autoref{tab:alpasim} compares \argus with camera planners on AlpaSim, NavSafe-$\infty$, and Odyssey, and \autoref{tab:hugsim} on HUGSIM.
The default \argus is smaller than every camera planner we evaluate, with 31.7M parameters, and is aligned on 25\,h of reconstructed camera views, against 1{,}800\,h of logged driving for TransFuser-K1 and 80k\,h for Alpamayo~1, a vision-language model about 330 times its size.
Yet it drives better than every camera planner on AlpaSim, HUGSIM, and NavSafe-$\infty$, and on Odyssey in non-reactive traffic, second only to SafeDrive~\cite{kim2026safedrive} in reactive traffic.
On the two tracks of AlpaSim (\autoref{tab:alpasim}a), it leads the best camera planner by 5.8 and 4.9 points, and Alpamayo~1 by 28.2 on PAI.
\argusplus, with a larger image backbone, reaches 87.3 on PAI and 96.1 on nuPlan, 78.9 on NavSafe-$\infty$, and 49.0 and 53.4 on Odyssey, where it also leads in reactive traffic: the recipe scales with the image encoder, though not on HUGSIM, where it scores 52.2.
Its lead grows where driving is hardest: on HUGSIM (\autoref{tab:hugsim}), its HD-Score is 48\% above that of the best camera planner, 53.6 against 36.1, a lead that holds under the realized HD-Score and at 10\,Hz (\autoref{sec:more_results:compare}), and on the safety-critical scenarios of NavSafe-$\infty$ (\autoref{tab:alpasim}b), its success rate is 11.1 points higher, 72.9 against 61.8.
Driving learned by RL in simulation thus carries over to camera images, making up for a smaller model and less camera data.

\vspace{0.4em}%
\noindent\textbf{It also outperforms prior self-play camera policies.}
\phantomsection\label{sec:exp:published}
\autoref{tab:positioning} lists the methods that bring a self-play policy to cameras: Gigapixel~\cite{rowe2026gigapixel}, DriveZero~\cite{he2026drivezero}, OPTED~\cite{dacol2026opted}, and TerraTransfer~\cite{xiong2026terratransfer}.
All four start from a self-play policy that reads a privileged vectorized state, while \argus starts from one that already drives from rasterized camera views, whose driving head relies on visual cues that real images also show.
None of them releases a trained camera policy, so we run \argus under the protocol of each and compare with the numbers they publish.
On the PAI track of AlpaSim, OPTED, which trains LTFv6 with DAgger from its self-play teacher, reports 41.8, against 84.2 for \argus (\autoref{tab:alpasim}a), though scored before AlpaSim's corridor fix, so the two are not strictly comparable.
Gigapixel and DriveZero report the official HD-Score on the 345 scenarios of HUGSIM's pre-challenge release in its default setting (\autoref{sec:protocol:hugsim} details how we run this simulator).
Under this protocol, \argus reaches 54.1, against 46.6 for DriveZero-Scale and 38.5 for Gigapixel (\autoref{tab:hugsim_bc}b).
TerraTransfer reports its realized HD-Score (\autoref{sec:protocol:benchmarks}), which scores the path that the ego drove, on the 88 scenarios from nuScenes in the non-default 10\,Hz setting.
Under this protocol, \argus reaches 62.1, against 49.0 for TerraTransfer (\autoref{tab:hugsim_bc}c).
Across these protocols, \argus leads every method that aligns or distills a vectorized policy, which supports that a raster policy is easier to bring to real images.

\subsection{Building \argusraster}
\label{sec:exp:pipeline}
\begin{table}[t]
  \centering
  \caption{\textbf{From self-play to \argusraster}, on raster views at 10\,Hz: AlpaSim scene score (PAI, nuPlan), HUGSIM realized HD-Score; best camera planner: TransFuser-K1, DrivoR, LTFv6 (Tabs.~\ref{tab:alpasim}, \ref{tab:hugsim_10hz}).}%
  \label{tab:pipeline}
  \footnotesize
  \setlength{\tabcolsep}{2.2pt}
  \begin{tabular}{lccc}
    \toprule
    Policy & PAI & nuPlan & HUGSIM \\
    \midrule
    {\color{black!60}Best camera planner} & {\color{black!60}78.4} & {\color{black!60}91.0} & {\color{black!60}51.3} \\
    \midrule
    Self-play, w/~action & 61.1 & 83.9 & 75.7 \\
    + logged traffic, w/~action & 89.3 & 96.2 & 71.2 \\
    \rowcolor{argusrow} + traj.\ head: \argusraster & 88.8 & 95.8 & 71.8 \\%
    \bottomrule
  \end{tabular}
\end{table}

\noindent\textbf{Self-play alone already drives well.}
Without a single driving log, the self-play-only policy already competes with planners trained on large driving corpora (\autoref{tab:pipeline}).
Driving with actions on raster views, with every simulator at 10\,Hz, it scores 61.1 on PAI, above the 56.0 of Alpamayo~1, a vision-language model of 10.5B parameters.
On HUGSIM, its realized HD-Score of 75.7 tops every policy we evaluate, far above the 51.3 of the best camera planner (Tabs.~\ref{tab:pipeline} and~\ref{tab:hugsim_10hz}).
Lastly, we compare it with its closest counterpart, \alberti, the driving policy that \pictura trains by self-play~\cite{yin2026pictura}, in the protocol of \pictura on the CARLA maps.
It is far safer: its collision rate is 0.002 against 0.060, and it drives 106\,km per infraction against 12, though it reaches fewer goals (\autoref{tab:selfplay_pictura} in \autoref{sec:more_results:selfplay}).

\vspace{0.4em}%
\noindent\textbf{Extending self-play to logged traffic closes the gap.}
On AlpaSim, however, the self-play-only policy trails the best camera planners, by 17.3 points on PAI and 7.1 on nuPlan (\autoref{tab:pipeline}).
AlpaSim differs from self-play in two ways.
Its cars replay their logs and do not slow down when the ego brakes, so a cautious ego is hit from behind, a known log-replay artifact~\cite{jaeger2025carl,charraut2025vmax}.
Its logs also include highway drives, faster than self-play's towns and cities allow.
Extending training with a mix of self-play and single-agent logged scenes for 342M steps, under 1\% of its 60B training steps, raises its score to 89.3 on PAI and 96.2 on nuPlan, above every camera planner, at a cost of 4.5 points on HUGSIM, where its 71.2 stays well above the best planner's 51.3.

\vspace{0.4em}%
\noindent\textbf{Trajectory head drives as well as actions.}
The trajectory head, trained last on the frozen network, produces trajectories, as end-to-end planners commonly do.
Using it to drive (last row of \autoref{tab:pipeline}), \argusraster stays within 0.6 points of its scores with actions on every benchmark: turning the policy into a planner costs no driving quality.

\subsection{Rendered-camera versus vectorized policy}
\label{sec:exp:vectorized}

\begin{figure}[t]
  \centering
  \resizebox{\linewidth}{!}{%
\figfont
\includegraphics{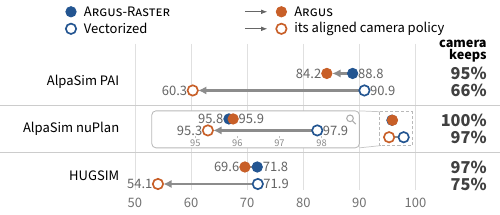}}
  \caption{\textbf{Policy swap}: score of each policy learned in simulation (blue) and of the camera policy aligned from it (orange), with the share kept on the right.
    Filled: \argusraster, on rasters, and \argus, on AlpaSim, and on HUGSIM at 10\,Hz; hollow: \vectorized, on a vectorized state, and its camera policy.}
  \label{fig:policy_swap}
\end{figure}

Most driving self-play policies~\cite{cusumanotowner2025gigaflow,cornelisse2025pufferdrive} do not observe the scene with visual cues: they read a vectorized state that the simulator gives them.
We train such a policy, \vectorized, with the same RL recipe in self-play as \argusraster (\autoref{sec:protocol:compare}).
\autoref{fig:policy_swap} follows each of the two policies from simulation (blue) to the camera policy aligned from it with the same losses and data (orange); as \vectorized has no scene latents, the latent loss of its camera policy targets the pooled features of its observed elements.
With this privileged input, \vectorized drives on par with \argusraster (blue): 90.9 against 88.8 on PAI, and 71.9 against 71.8 on HUGSIM.
\argusraster, however, learns to drive from the same visual cues that the camera policy later receives, while \vectorized relies on a state that cameras do not give, asking the alignment to act on information it cannot observe~\cite{nguyen2026lead}.
As a result, the camera policy aligned from \argusraster keeps a larger share of its score (orange in \autoref{fig:policy_swap}): 95\% on PAI and 97\% on HUGSIM, against 66\% and 75\% for the one aligned from \vectorized.

\subsection{Ablations}
\label{sec:exp:ablation}
\phantomsection\label{sec:exp:alignment}

We ablate the trajectory head and the alignment (\autoref{sec:more_results:ablation}).

\begin{table}[t]
  \centering
  \caption{\textbf{Ablations}: AlpaSim scene score (PAI, nuPlan) and HUGSIM realized HD-Score in its default setting of \argus, as mean and standard deviation over three seeds of the alignment; see \autoref{sec:more_results:ablation} for each setting. Shaded: default.}
  \label{tab:alignment_losses}\label{tab:traj_head}
  \vspace{2pt}%
  \ifdefined\ablboxa\else\newsavebox\ablboxa\newsavebox\ablboxb\newdimen\ablwmax\newdimen\ablwa\newdimen\ablwb\fi
  \sbox\ablboxa{  \scriptsize
  \setlength{\tabcolsep}{1.6pt}
  \renewcommand{\arraystretch}{1.2}%
  \begin{tabular}{llccc}
    \toprule
    Head & Training & PAI & nuPlan & HUGSIM \\
    \midrule
    \rowcolor{argusrow} Flow & DAgger & \seedcell{84.21,83.30,84.94} & \seedcell{95.85,95.96,95.99} & \seedcell{57.4,56.4,56.4} \\
    Flow & BC & \seedcell{82.00,83.42,80.77} & \seedcell{95.63,96.41,95.95} & \seedcell{55.9,54.2,55.4} \\
    \midrule
    MSE & DAgger & \seedcell{82.11,81.46,81.39} & \seedcell{95.48,95.17,95.59} & \seedcell{54.1,50.6,53.1} \\
    WTA & DAgger & \seedcell{76.69,75.82,76.29} & \seedcell{95.22,95.48,95.53} & \seedcell{52.4,51.7,51.5} \\
    \bottomrule
  \end{tabular}
}\sbox\ablboxb{  \scriptsize
  \setlength{\tabcolsep}{1.6pt}
  \begin{tabular}{lccc}
    \toprule
    Losses & PAI & nuPlan & HUGSIM \\
    \midrule
    \rowcolor{argusrow} $\mathcal{L}_\text{\argus}$ & \seedcell{84.21,83.30,84.94} & \seedcell{95.85,95.96,95.99} & \seedcell{57.4,56.4,56.4} \\
    \midrule
    w/o $\mathcal{L}_\text{KL}$ & \seedcell{82.46,82.41,84.40} & \seedcell{95.48,95.30,95.38} & \seedcell{60.4,59.4,57.5} \\
    w/o $\mathcal{L}_V$ & \seedcell{85.22,83.89,84.40} & \seedcell{96.09,95.62,95.80} & \seedcell{56.9,55.3,56.1} \\
    w/o $\mathcal{L}_\text{lat}$ & \seedcell{68.57,63.11,73.12} & \seedcell{94.16,92.99,95.14} & \seedcell{47.0,39.6,47.3} \\
    $\mathcal{L}_\text{lat}$ only & \seedcell{84.94,83.87,83.09} & \seedcell{95.33,95.51,95.23} & \seedcell{60.1,58.3,59.0} \\
    \bottomrule
  \end{tabular}
}%
  \ablwmax=\ifdim\wd\ablboxa>\wd\ablboxb \wd\ablboxa\else\wd\ablboxb\fi
  \ablwa=\dimexpr 0.495\columnwidth*\wd\ablboxa/\ablwmax\relax
  \ablwb=\dimexpr 0.495\columnwidth*\wd\ablboxb/\ablwmax\relax
  \begin{minipage}[t]{0.495\linewidth}\vspace{0pt}\centering
  {\scriptsize\bfseries (a) Trajectory head}\par\vspace{2pt}
  \resizebox{\ablwa}{!}{\usebox\ablboxa}
  \end{minipage}\hfill
  \begin{minipage}[t]{0.495\linewidth}\vspace{0pt}\centering
  {\scriptsize\bfseries (b) Alignment losses}\par\vspace{2pt}%
  \resizebox{\ablwb}{!}{\usebox\ablboxb}
  \end{minipage}
\end{table}

\vspace{0.4em}%
\noindent\textbf{Trajectory head.}
The trajectory head learns by DAgger: it drives during its own training, so it also learns from the states its own plans lead to (\autoref{sec:method:traj}).
Against behavior cloning, where it learns only from the states of the action head (\autoref{tab:traj_head}a), DAgger pays off where errors compound: on the long scenes of PAI, 84.2 against 82.1, and on the harder scenarios of HUGSIM, 56.7 against 55.2, while the two are level on the short scenes of nuPlan, 95.9 and 96.0.
The flow head also beats two MLP heads trained the same way, 81.7 and 76.3 on PAI: one that regresses a single plan with an L2 loss (MSE), and one that handles multi-modality by predicting 8 plans, trained winner-takes-all (WTA), as in DrivoR~\cite{kirby2026drivor}.

\begin{figure}[t]
  \centering
  \input{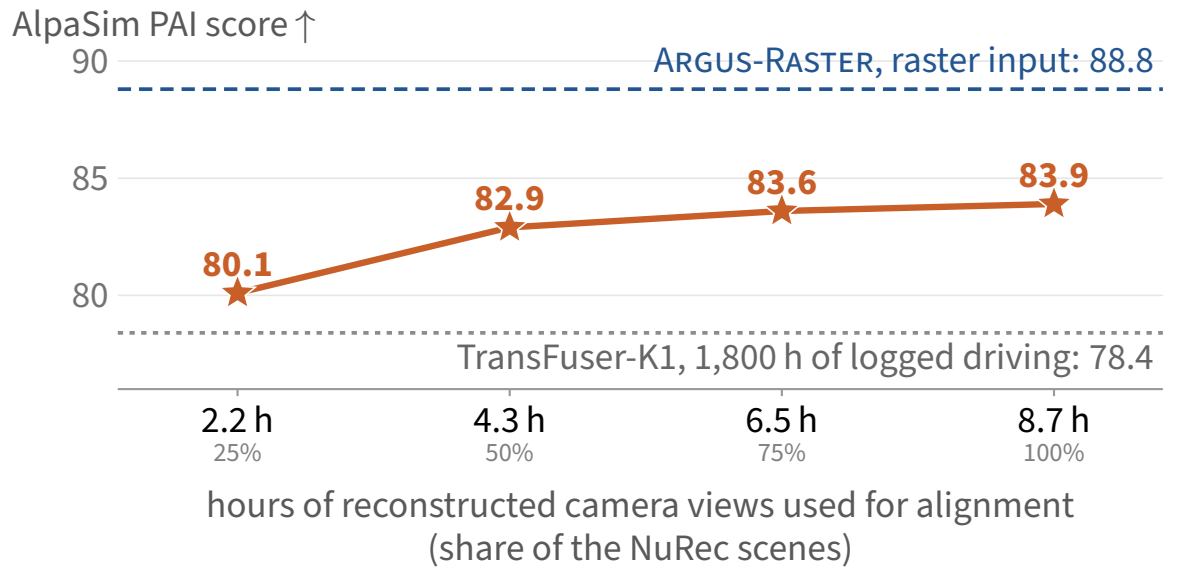}
  {\figfont\dataamountplot{(25,80.1) (50,82.9) (75,83.6) (100,83.9)}{25/80.1,50/82.9,75/83.6,100/83.9}{88.8}{76}{80,85,90}{%
    \draw[black!45, dash pattern=on 1pt off 1.2pt, line width=0.7pt] (14,78.4) -- (110,78.4);
    \node[anchor=north east, text=black!60, inner sep=1.5pt] at (110,78.4) {TransFuser-K1, 1{,}800\,h of logged driving: 78.4};}}
  \vspace{-0.5em}
  \caption{\textbf{Amount of alignment data}: AlpaSim PAI score of \argus aligned on subsets of the 1{,}579 NuRec scenes completed by \argusraster, for the same number of steps.
    Dashed: \argusraster on rasters; dotted: the best camera planner.}
  \label{fig:data_amount}
\end{figure}
\vspace{0.4em}%
\noindent\textbf{Alignment losses.}
The alignment of \argus simply sums three standard losses of prior alignment and distillation methods, with equal weights (\autoref{sec:method:alignment}).
\autoref{tab:alignment_losses}b removes each in turn, and no setting is best everywhere: without $\mathcal{L}_V$ on PAI (84.5), without $\mathcal{L}_\text{KL}$ on HUGSIM (59.1), and all three on nuPlan (95.9).
The latent loss alone is a strong candidate, on par with all three on PAI (84.0 against 84.2) and above them on HUGSIM (59.1 against 56.7), while removing it costs 15.9 points on PAI.
This further supports designing the self-play policy for alignment: since \argusraster drives from the same visual cues as the camera policy, matching its latents carries most of its driving.
Every combination with the latent loss beats all camera planners on the three benchmarks (Tabs.~\ref{tab:alpasim} and~\ref{tab:hugsim_4hz}): the alignment is robust to its losses, and weighting them may help further.

\vspace{0.4em}%
\noindent\textbf{Alignment data.}
\autoref{fig:data_amount} varies the amount of camera views used for the alignment, and shows that little is needed: a quarter of our 1{,}579 collected NuRec scenes (\autoref{sec:impl:align_data}), 2.2\,h of reconstructed views, already scores 80.1 on PAI, above the 78.4 of the best camera planner, TransFuser-K1, which imitates 1{,}800\,h of logged driving.
Half of the scenes score 82.9, within a point of the 83.9 reached with all of them.
\argus thus needs little camera data: its driving comes from 5M hours of RL in simulation, and compute spent there, not camera hours, is what buys its performance.

\section{Conclusion}
\label{sec:conclusion}

\argus shows that a policy learned by RL on rasterized camera views transfers to realistic camera images by aligning only its image encoder.
Its driving head relies on visual cues that a pretrained image encoder also finds in real images, so a few hours of reconstructed camera views suffice, where imitation-learned planners need thousands of hours of logged driving.
The camera policy keeps most of the driving of the raster policy, far more than one aligned from a vectorized policy, and outperforms in closed loop every imitation-learned camera planner we evaluate, everywhere but on Odyssey in reactive traffic.
Driving skills can thus be bought with simulation compute.
A larger image backbone already closes that gap and raises the PAI score from 84.2 to 87.3, though not the HUGSIM score, so stronger image encoders should carry more of the driving of the raster policy.

\paragraph{Limitations.}

The raster twins need a few hours of 3D boxes, maps, and traffic-light states, which can be noisy, for the alignment. Our camera views are reconstructed: aligning on the logged frames also drives well, and evaluating on deployed camera data remains open.

\section*{Acknowledgments}
We thank Ellington Kirby, whose earlier work on the vector simulator laid the groundwork for this project, and Lan Feng for generous discussions and early experiments on raster twins.
We acknowledge the EuroHPC Joint Undertaking for awarding this project access to the EuroHPC supercomputer LEONARDO, hosted by CINECA (Italy) and the LEONARDO consortium through EuroHPC AI Factory access calls.
This work was granted access to the HPC resources of IDRIS under the allocations A0201016203, A0211018323, and AD011018323R1 made by GENCI.
We acknowledge the EuroHPC Joint Undertaking for awarding this project access to the EuroHPC supercomputer JUPITER, hosted by the J\"ulich Supercomputing Centre (Germany), through a EuroHPC Regular Access call (project EHPC-REG-2026R01-243).
We acknowledge the EuroHPC Joint Undertaking for awarding this project access to the EuroHPC supercomputer Karolina, hosted by IT4Innovations (Czech Republic), through a EuroHPC Development Access call (project EU2026D03-020).
This work was supported by the VISA DEEP AI chair, funded by the Agence Nationale de la Recherche (ANR, ANR-20-CHIA-0022), and by the PostGenAI@Paris cluster, which benefited from government funding managed by the ANR under the France 2030 program (ANR-23-IACL-0007).

{
    \small
    \bibliographystyle{ieeenat_fullname}
    \bibliography{main}
}

\input{sec/X_appendix}

\end{document}